# Use HMM and KNN for classifying corneal data

Payam Porkar Rezaeiye
* department of computer, damavand branch, islamic azad university, Damavand, iran, Email: porkar@damavandiau.ac.ir

mehrnoosh bazrafkan Zarghan Branch, Islamic Azad University, Zarghan, Iran, Email: mehrnoosh.bazrafkan@gmail.com

ali akbar movassagh Department of Electrical Engineering _ Sharif University of Technology_ Tehran_Iran, Email: a.movassagh@gmail.com

Mojtaba Sedigh Fazli department of computer, damavand branch, islamic azad university, Damavand, iran, Email: Mojtabafazli@yahoo.com

Gholam hossein bazyari varamin University of Science and technology, Varamin, Iran, Email: bazyari@gmail.com

*Abstract—* **These days to gain classification system with high accuracy that can classify complicated pattern are so useful in medicine and industry. In this article a process for getting the best classifier for Lasik data is suggested. However at first it's been tried to find the best line and curve by this classifier in order to gain classifier fitting, and in the end by using the Markov method a classifier for topographies is gained.**

**What are mentioned in this article are supposed to gain a strong classifier so that under Marko theory can choose eyes appropriate for corneal graft.**

*1. Introduction:*

These days one of the most important issues that is useful in many usages, is to find the best fitting line and curve among the data. In fact the aim of this job is to find limits for every category of data and by this we can find the order of data and do the classification with high accuracy. Using the fitting line and curve has a vast usages for instance one of the new usages of the best curve is in quantity analyze of data, in other words we are this method to determine the stickiness of proteins of DNA. [1] In other article there's a hint for the usage of artificial nervous system for dispersal data, this method is useful in data fitting [2] one of the other usages of the fitting curve is to determine the human height's growth [3] and also in other article the best fitting curve is used to find time series [4].

*2. Finding the best fitting line and curve by Markov method:*

This is not the only aim in this article to find the fitting line and point by Markov method but it's been used for creating a down to up logic for designing a particular classifier and in fact finding the fitting line and curve is the beginning of finding a strong classifier by Markov chains.

*2-1. Finding the best line, which has the closest distance to all the points [5]:*

We presume, we have m as (xi, yj), i=1… m. For finding the best line we need to find the best a,b in the function below:

$$F(x) = ax+b$$

In figure 1 part A there is a sample for the best line to some points. In order to find the best line we need to calculate the y distance of each point from the curve and the curve which has the minimum sum of distance is the best one, these distances are shown in figure 1 part B.

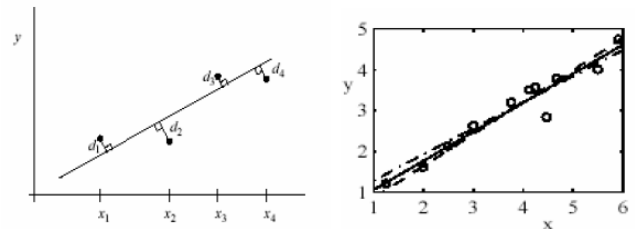

Figure1. A. Sample of line fitting.  B. distance of points to best line

In designing classifier, in order to gain the best line by using Markov models for each stage we use the line gained in the previous stage by Markov, In other words each stage has the information of that stage and the previous stage (the main feature of Markov chain), in a way that in the first stage we fit the line to two points and then in the next stage we fit the third point of the other line by the gained line which has the less distance from each three points in the sheet and in this way we continue the processes to find the line that fits to all the points in the sheet. There is a sample of results for four points in the figure 2.

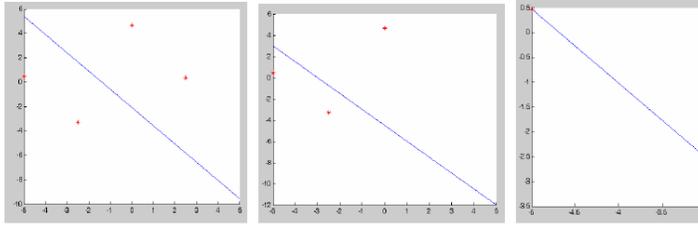

Figure2. Line fitting with HMM (A. stage 2 B. stage 3 C. stage 4)

## 2-2. Finding the best curve, closest to all points:

To find this we find the distance of the three first points to the curve $y=ax^2+bx+c$ and then we look for the best quadratic curve which has the less distance from these three points. After finding that particular curve we go through next stage and repeat all the stages above again based on the curve we've found in the previous stage and also in order to get the optimum result we use Viterbi algorithm (This algorithm helps us to find the best curve in each stage according to the back ground of previous curves and function that to the new set of point). Figure 3 show the two stage of curve fitting.

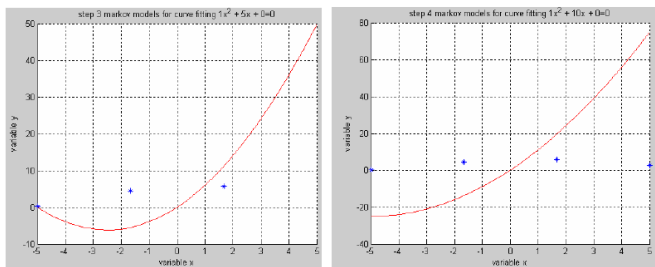

Figure3. Curve fitting with HMM (A. stage 3 B. stage 4)

In resumption we'd express an abstract on medical important issues in diagnosing corneal and then explain the Markov model which was used.

## 3. Corneal Topography

A significant issue in this article is using a statistic method for classifying different corneal in order to improve corneal graft, in this part we'll discuss the issue. One of the basic problems in graft surgery is to examine the quality of the dedicated corneal, because if the quality is low, corneal graft won't have a significant effect on the sight of the person, for instance a factor which is very important here is not to have any previous surgery on eyes and specially Lasik surgery besides not having illnesses like corneal hump are the parameters which are important for corneal gets grafted.

One of the important problems in this project is to lose eye pressure after death which has a great effect on topography. Usually after death all parts of the body like eye get swelled which has an effect on the thickness of corneal and the topography so when the system gives us the results, they would have some problem, however by the outcome preference the accuracy will increases.

## 3-1. Medical scrutiny of the different pictures from the systems:

First its better get familiar by different types of pictures related to corneal that physicians deal with. The most important picture is the one used for scrutiny the surface of the corneal called axial map or pathfinder, this picture in general shows as the thickness of different parks of the corneal and in this task there had been more work on these two pictures to result preferences [7]. Also Keratometry view picture gives some information related to the thickness of different parts of the corneal, the important point here is that in this picture all three circles and their thickness are important for graft surgery but for Lasik surgery only two of them are considered [7]. Other types of the pictures is numerical view which shows the density of different parts thickness and we can realize how good the corneal is for the surgery by using this picture, to some extent [7]. We can't use photo Keratoscope for diagnosing Lasik but sever astigmatism and any surgery on corneal would be obvious and the effect is shown some circles gathered in the direction of one axis. The final type of picture we discuss here is profile view which can be useful for diagnosing as it shows corneal's curve beside the thickness [7] in figure 4 some types of the pictures are shown.

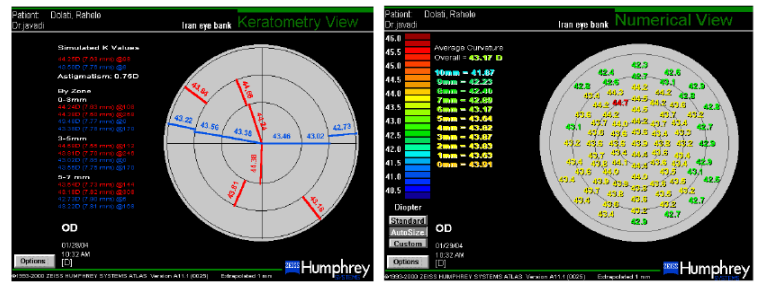

A                B

Figure4. A. Numerical View   B. Keratometry View

## 3-2: Discussing the suggested method by the physician:

In figure 5 there samples of topography of a healthy corneal, as we see different preferences are upper in both pictures which are accurate for the samples of one class.

In general the thickness of the corneal is usually 420 and 800 ml. but those which had a Lasik surgery have lower thickness also the topography of the person with hump in corneal won't be symmetrical.

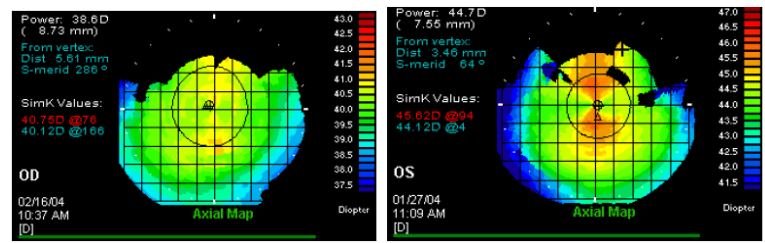

Figure5. Sample of healthy corneal topography.

So we can use these two factors to realize the quality offered corneal. The important point, results from the two above pictures is that although both eyes are healthy but the night one has lower quality compare to the left one, that's because the night eye has some astigmatism on its 90 axis however as we see in the pictures the left one has a steady surface so has a better quality for the graft surgery (for better thickness). Figure 6 refers to two topography of corneal, introduces by normal system, but according to specialists only the left corneal is fine for the surgery. We have to consider, although the quantities of the figure don't contain natural quantities of a normal and fine corneal but because of the symmetrical exist on the up and down of the center of the corneal, this corneal is fine for graft surgery, however the receiver of this corneal needs to wear glasses after the surgery. But the right corneal is not appropriate for graft surgery and it's because as we see in the pictures around the center of the picture on the upper side of the corneal is thicker and the down side has less curve (the quantities are comparable and calculable according to colors), the above corneal has some problems at the upper side according to the specialists which might be caused by Lasik surgery and etc. so in general as we face graft surgery subject the important issue is the symmetry of the corneal and also it's appropriate thickness and if want to have accurate quantities for the depute of the offered corneal we have to use this topography picture, in resumption a process for concluding two preferences; thickness and symmetrical, has been discussed which has been under the function of the classifying system.

### *4. Preprocessing:*

For doing this project we have used collected images from two devices Humphrey, Orbscan [8]. Although corneal topography from these two devices was different, we use image processing algorithms over these images to obtain additional information in a way appropriate feature extraction. We also show a high precision classifier. In next sub section we have represented all related image processing algorithms and a description of two devices.

### *4.1 preprocessing for Humphrey device:*

This device has nine output images types that have different applications. Here we have used from topography images which named Pathfinder. This step concerned on all types of input images. For instance output images of Humphrey device that is shown in Figure 5. As figure 5 shows the image of corneal topography have surplus details. Therefore in first step we have needed to divide sections. For this reason we use a preprocess program to extract 159*159 dimensions square corneal topography from input machine images. The result has shown in Figure 6.

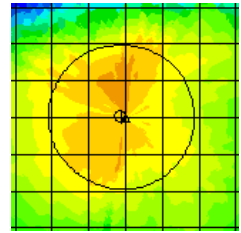

Figure6. Division section of corneal images in dimensions of 159*159

As we can see in figure 6, this image composed of lines to help ophthalmologist. These lines sometimes have misunderstandings for machines. So we should use an elimination algorithm before every evaluation. To gain this purpose we have used image processing methods with the help of average neighbor color to find filled black point. Also we have used median filter. To obtain the best result, we have done a little change in popular median filter that has an ordering color capability. Darker colors more than thresholds were eliminated from list and median color have gotten from primary color image. The sample of this operation presented in Figure 7.

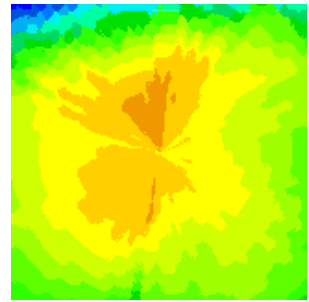

Figure7. The image obtained from eliminated lines program.

As we can figure out from the Figure 7, distinct this program retrieved principle images very suitable. We should consider that this filter only operates in black points and other information remains without any changes. It has a wide application where we want to eliminate all black points and where ever we cannot obtain real data and also does not have any changes to real data. In figure 8 measurement associations of red greens and blue colors are shown.

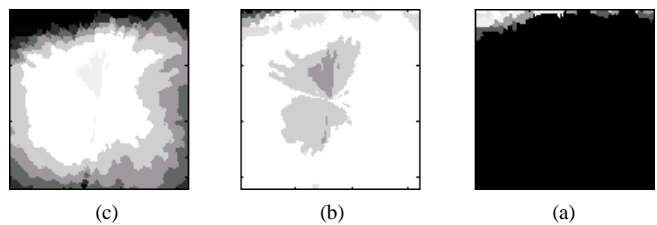

(c)  (b)  (a)

Figure8. The image shows the role of three principle colors that have converted to contrast image (a) blue (b) green (c) red.

Figure 9 shows final image after using of filtering process.

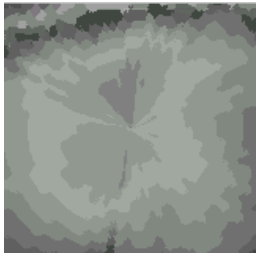

Figure9. The final image obtained after filtering process.

### 4-2. Obtaining Features:

Obtaining a feature is one of the most important processes in every recognizing system, and it's very important to supply features that can distinguish data very well. Here both mathematically and mentally. Sensitively features are used for creating separating space. In this task a lot of features are discussed to obtain desired quantities that distribute data in a sample space and among them all we can mention issues such as correlation, auto correlation, cross correlation, fast furia transform and spectrum of the discussed picture, all of them are mathematical features. Based on this, some features like picture spectrum are not appropriate for classifying and other features such as correlation, auto correlation, cross correlation and furia transform have their own effect on distinguishing classes of Lasiked-eyes and not Lasiked-eyes, so using one of them would be enough. The distributions of patterns are too complicated and using one feature doesn't allow distinguishing the results well. Despite the FFT is not a good feature by itself but it's been considered because it shows good results in the mixture by correlation in a way that it was possible to distinguish classes. On the other hand as the patterns are gathered together in a complicated manner and besides the number of sample patterns were few so to increase the insurance for better distinguishing the classes, searching other features seems to be necessary. Also according to the information gained by the second system or Orbscan, it was detected that this system does other measurements too which are indistinguishable in the discussed pictures. For instance of one of the pictures (specially shows the thickness of the corneal and called Pochemtery) different quantities obtained by the system showed that these quantities are (5) which shows the maximum thickness of the five main points of the corneal.

With other comparison some more features were obtained as below:

1: (Maximum Power) maximum value = middle value of the corneal − maximum value of the four sides numbers.

2: (Minimum Power) minimum value = middle value of the corneal − minimum value of the four sides number

### 4-3. Statistical classifier of Nearest Neighbor:

In KNN rule for classifying each unknown sample of x in a n dimension space K finds the nearest neighbor base on a distance scale and among this, K finds nearest neighbor of the class contains majority and attribute x to that class. In order to find K nearest neighbor x starts searching in a long distance from that and in this way there'll be the case of uncertainness in decision making. Here for decreasing the fault in decision making, making decision would be rejected or called the sample x reject.

### 4-4. Using Markov in recognizing the quality of the offered corneal:

As mentioned before, the first step is to find the best line and curve in order to find fine ideas to straighten the classifier that must be used in this stage. We used Markov method to examine the quality of offered corneal related to the designed system. But before we can use this method there is a pre-process stage in which we change the out coming pictures from the system to the fine ones for making decisions, like omitting window shape lines and removing colors in the picture. (in making pictures black and white the information won't get lost) but before discussing the designed model it's better to have an abstract on Markov models and the previous tasks similar to this project. In more than one decade hidden Markov models were used in recognizing speech. [12] And in recent years they are used more in recognizing behavior (face) and in general recognizing patterns.

Only a few attempted has been made in order to use HMM in recognizing patterns like using this method in identifying facer. [11, 13], in general HMM concentrate on series of one dimension signals. While a normal picture is presented by a typical two dimension matrix. This problem can be solved by using a slipper window that covers the width of the picture and moves from up to down. The quantities of reflecting window would cross the HMM process. Frequent window are put on each other in order to avoid cutting important features and the lost information would be gained frequently from feature axes. Here we've used the symmetry in the pictures and also the corneal thickness as main feature and correlation as the secondary feature for instructing the system. There some algorithm available for both instruction and testing HMMs. The aim of instructing the HMM is to extend the parts it covers. As the possibilities improve the real frequent of S position can be determined to observe x. Any way there should be enough instruction data so that internal statistic models can be made. Improving and convergence of the process of determining Baum-Welch can find the local minimum of the HMM.

Parameters for a know group of instruction data. [9]

In a reasonable manner giving quantities to the estimates can help to find general optimum solutions. For testing and recognizing the Viterbi algorithm determines the frequency of S position with the possibility given above and in this way a special series of frequent observance of X reveals. (To maximize p(s|x,x). in this part we gave a simple explanation

of algorithm used for instruction and testing hidden Markov model. The first algorithm is the Baum-Welch method which uses one estimate of the parameters of the model. The second algorithm is the Viterbi method which valuate the ability of a HMM in the explanation of a special observance for more details, refer to [12]. Finally in order to show the power of designed HMM we'd compared that to a statistical classifier called closest neighbor classifier. The results in table1 show the excellence of it and these classes are easily separated.

| Model | Correlation & FFT & Max Diff | Correlation & FFT & Min Diff | Correlation & FFT & Max-Min Diff |
|---|---|---|---|
| KNN | %65 | %70 | %75 |
| HMM | %71 | %77 | %86 |

Table1. Compare between HMM & KNN

*5- Conclusion:*

In this article we mentioned a usage of HMM in finding the best fitting line and curve which is very useful indifferent fields nowadays. Then the problems and issues we deal related to corneal was expressed and then we discussed corneal classification using a statistical method called HMM (hidden Markov model) and by this method of learning machine we gained fine results for determine for the qualities of offered corneal. However this usage can be improved by finding other features, also comparisons have been made between the suggested method and other classifier called closest neighbor which shows the excellent ability of HMM in this pattern and method. For more we can use intelligent methods to determine the scale of astigmatism and also far-sightedness and short-sightedness in a person by using a topography picture.

In this paper we have introduced and compared HMM and KNN method for classification [18, 19]. Also appropriate feature extraction could be used to improve speed and recognition ratio of classification. And we have showed that we have taken better effectiveness. Our plan is to introduce a more effective algorithm with the help of Support Vector Machine.

REFERENCES

[1] Susan E.Shadle , Douglas F.Allen , Hong Guo , Wendy K. Pogozelski , John S. Bashkin and Thomas D. Tullius ' quantitative analysis of electrophoresis data : novel curve fitting methodology and its application to the determination of a protein-DNA binding constant ' john Hopkins university , 1996.

[2] Miklos Hoffmann ' scattered data fitting by artificial neural network ' university of eger, 2002.

[3] Hosein Hasani , Mohamad Zokaei , Ali Amidi ' a new approach to polynomial regression an dits application to physical growth of human height ' shahid beheshti university ,2002.

[4] Malcom R.Forster ' time series and curve–fitting:how are they related? 'January 1999.

[5] Gerald Recktenwald ' least squares fitting of data to a curve ' Portland university November 2001.

[6] Thomas Mckay ' a clinical guide to the Humphrey corneal topography system ' zeizz publisher, 1998.

[7] 'humphret atlas corneal topography system with mastervue software', 1998.

[8] Noor Clinic, http://www.noorvision.com, November, 2004.

[9] K. Lua "Hidden markov model-HMM", lecture node in HMM, 1999.

[10] Donald. Tanguay "Hidden Markov models for gesture recognition" 1995.

[11] MIT Lecture node ' hidden markov modeling ' lecture 10 session2003.

[12] B.Asherman , H.Bunke ' combination of face classification for perso identification ' proceeding of the 13th IAPR international conference on pattern recognition(ICPR),1996.

[13] X.Haung, Y.Ariki, M.Jack ' hidden markov models for speech recognition ' Edinburg univ.press, 1990

[14] C.Nyffenegger 'gesichtserkennung mit hidden markov modellen ' Master's Thesis, IAM, bern university, 1995.

[15] L.R.Rabiner ,B.H.Juang ' an introduction to hidden markov models' IEEE ASSP magazine ,1986.

[16] F.Samaria, S. Young "HMM-Based Architecture for face identification ", Image and Vision Computing, `1994.

[17] T.starner, A. Pentland, "visual Recognition of American sign Language Using Hidden Markov Models", Proceeding of International workshop on Automatic Face and Gesture Recognition IWAFGR, 1995.

[18] Payam Porkar Rezaeiye , Saeed Setayeshi "Medicine Irregular Pattern Recognition Using Statistical Method Based On Hidden Markov Model", Azad University Science & Research Branch Artificial Intelligence & Robotic Engineering Department, November 2006.

[19] Payam Porkar Rezaeiye, Pasha Porkar Rezaeiye, Elahe Karbalayi, Mehdi Gheisari "Statistical method used for doing better corneal junction operation" , International Conference on Electronics, Nanomaterials and Component (ICENC 2012), Chengdu, China, May 5-6, 2012. (Index in SCOPUS, CPX, CSA, CA, ISI, ISTP, CPCI, Web of Science, IEE).